\documentclass{article} 

\usepackage[T1]{fontenc}
\usepackage{ifthen}
\usepackage{mdwlist}
\usepackage{amsmath,amssymb,amsfonts,amsthm}
\usepackage{bm}
\usepackage[colorlinks=true,linkcolor=blue,citecolor=blue,urlcolor=blue]{hyperref}
\usepackage{enumitem}
\usepackage{graphicx}
\usepackage{xspace}
\usepackage{verbatim}
\usepackage{algorithm}
\usepackage{algpseudocode}
\usepackage[margin=1in]{geometry}
\usepackage{color}
\usepackage{thm-restate}
\usepackage{latexsym}
\usepackage{epsfig}
\usepackage{multirow}
\usepackage{xcolor,colortbl}
\usepackage[colorinlistoftodos,textsize=scriptsize]{todonotes}
\usepackage[round]{natbib}
\usepackage{tablefootnote}
\usepackage{subcaption}
\def\withcolors{1}
\def\withnotes{1}

\renewcommand{\epsilon}{\ve}
\def\ve{\varepsilon}

\newcommand{\pr}[2][]{\mathrm{Pr}\ifthenelse{\not\equal{}{#1}}{_{#1}}{}\!\left[#2\right]}
\newcommand{\R}{\mathbb{R}}

\newtheorem{theorem}{Theorem}

\newtheorem{remark}[theorem]{Remark}

\newtheorem{definition}[theorem]{Definition}

\numberwithin{theorem}{section} 
\numberwithin{nontheorem}{section} 
\numberwithin{proposition}{section} 
\numberwithin{observation}{section} 
\numberwithin{remark}{section} 
\numberwithin{fact}{section} 
\numberwithin{lemma}{section} 
\numberwithin{claim}{section} 
\numberwithin{corollary}{section} 
\numberwithin{case}{section} 
\numberwithin{dfn}{section} 
\numberwithin{definition}{section} 
\numberwithin{question}{section} 
\numberwithin{openquestion}{section} 
\numberwithin{res}{section}

\ifnum\withcolors=1
  \newcommand{\gcolor}[1]{{\color{red}#1}} 
\else

  \newcommand{\gcolor}[1]{{#1}}
\fi

\ifnum\withnotes=1
  \newcommand{\gnote}[1]{\par\gcolor{\textbf{G: }\sf #1}} 
  \newcommand{\gfootnote}[1]{\footnote{{\bf \gcolor{Gautam}}: {#1}}}

\else
  \newcommand{\gnote}[1]{}
  \newcommand{\gfootnote}[1]{}

\fi
\newcommand{\ignore}[1]{\leavevmode\unskip} 

\newcommand*\samethanks[1][\value{footnote}]{\footnotemark[#1]}

\title{Differentially Private Fine-tuning of Language Models\thanks{Aside from the first and second authors, all other authors are listed in alphabetical order.}}

\author{
 Da Yu\thanks{Sun Yat-sen University. {\tt yuda3@mail2.sysu.edu.cn}. Work was done while an intern at Microsoft Research Asia.}
\and 
Saurabh Naik \thanks{Microsoft. {\tt \{snaik, lukas.wutschitz\}@microsoft.com}.}
\and
Arturs Backurs \thanks{Microsoft Research. {\tt \{arturs.backurs, sigopi, huseyin.inan, jakul, amonteiroman, yekhanin\}@microsoft.com}.}
\and
Sivakanth Gopi\samethanks[4]
\and	
Huseyin A. Inan\samethanks[4]
\and
Gautam Kamath\thanks{Cheriton School of Computer Science, University of Waterloo. {\tt g@csail.mit.edu}. Supported by an NSERC Discovery Grant.}
\and
Janardhan Kulkarni\samethanks[4]
\and 
Yin Tat Lee\thanks{University of Washington and Microsoft Research. {\tt yintat@uw.edu}.}
\and
Andre Manoel\samethanks[4]
\and
Lukas  Wutschitz\samethanks[3]
\and
Sergey Yekhanin\samethanks[4]
\and
Huishuai Zhang\thanks{Microsoft Research Asia. {\tt huzhang@microsoft.com}.}
}

\begin{document}

\maketitle

\begin{abstract}
We give simpler, sparser, and faster algorithms for differentially private fine-tuning of large-scale pre-trained language models, which achieve the state-of-the-art privacy versus utility tradeoffs on many standard NLP tasks.
We propose a meta-framework for this problem, inspired by the recent success of highly parameter-efficient methods for fine-tuning. 
Our experiments show that differentially private adaptations of these approaches outperform previous private algorithms in three important dimensions: utility, privacy, and the computational and memory cost of private training. 
On many commonly studied datasets, the utility of private models approaches that of non-private models. 
For example, on the MNLI dataset we achieve an accuracy of $87.8\%$ using RoBERTa-Large and $83.5\%$ using RoBERTa-Base with a privacy budget of $\epsilon = 6.7$. 
In comparison, absent privacy constraints, RoBERTa-Large achieves an accuracy of $90.2\%$.
Our findings are similar for natural language generation tasks.
  Privately fine-tuning with DART, GPT-2-Small, GPT-2-Medium, GPT-2-Large, and GPT-2-XL achieve BLEU scores of 38.5, 42.0, 43.1, and 43.8 respectively (privacy budget of $\epsilon = 6.8,\delta=$ 1e-5) whereas the non-private baseline is $48.1$.
All our experiments suggest that larger models are better suited for private fine-tuning: while they are well known to achieve superior accuracy non-privately, we find that they also better maintain their accuracy when privacy is introduced. 
\end{abstract}

\section{Introduction}
Deep learning models are well known to leak sensitive information about the dataset when trained using conventional methods~\citep{ShokriSSS17, CarliniLEKS19, CarliniTWJHLRBSEOR21}. 
To combat this issue, models can instead be trained to guarantee differential privacy (DP)~\citep{DworkMNS06}, a strong notion of data privacy which limits the influence of any individual training point on the final model.
While DP is one of the few approaches capable of providing machine learning models with rigorous 
privacy guarantees, it generally comes at a cost in terms of test accuracy. 
One oft-cited explanation is that the constraint of DP necessitates much more training data~\citep{TramerB21, Feldman20, BrownBFST21}.
Unfortunately, more training data may be hard to acquire, particularly in settings where privacy is a concern. 

Parallel to these developments, Transformer-based~\citep{VaswaniSPUJGKP17} large language models (LLMs), including the BERT~\citep{DevlinCLT19, LiuOGDJCLLZS19} and GPT~\citep{RadfordNSS18, RadfordWCLAS19, BrownMRSKDNSSAAHKHCRZWWHCSLGCCBMRSA20} families, have enabled significant progress in natural language processing, achieving state-of-the-art accuracy in almost every task considered.
These models are first pre-trained on an extremely large and diverse public dataset.
The weights are then fine-tuned for each task of interest using a much smaller task-specific dataset.
For example, a single pre-trained GPT-family model may be fine-tuned for various downstream tasks, such as email reply suggestion, sentence completion in text editors, language translation, and more.
This two-stage paradigm can naturally be adapted to solve tasks in private learning, automatically addressing the aforementioned data shortage issue via the massive scale of the public pre-training dataset.
One may pre-train the model on public data as usual,\footnote{Despite the fact that the pre-training data is public, there may nonetheless be privacy concerns related to personal or copyrighted data. However, since these pre-trained models have already been released, any associated privacy loss has already been incurred.} but then fine-tune the model \emph{privately}.

Despite the success of these models, task-specific fine-tuning introduces a number of technical challenges.
In the non-private setting, the immense size of LLMs makes it impractical to fine-tune the full model and store a separate copy of the parameters for hundreds of downstream tasks.
Things only get worse with privacy, which leads to overheads in terms of running time, memory usage, and most importantly, accuracy. 
The magnitude of noise introduced to a model due to DP grows as the model size increases~\citep{BassilyST14, AbadiCGMMTZ16, BunUV14}, which can overwhelm any signal for larger models.
A recent line of work in the non-private literature has proposed parameter-efficient methods to alleviate the issues of storage and computational cost for fine-tuning~\citep{houlsby2019parameter, LiL21, AghajanyanZG20, hu2021lora, mahabadi2021compacter}.
The main focus of our work is to explore parameter-efficiency in the context of private learning.

\subsection{Our Contributions}
\begin{figure}[t]
\begin{centering}
\includegraphics[width=0.8\textwidth]{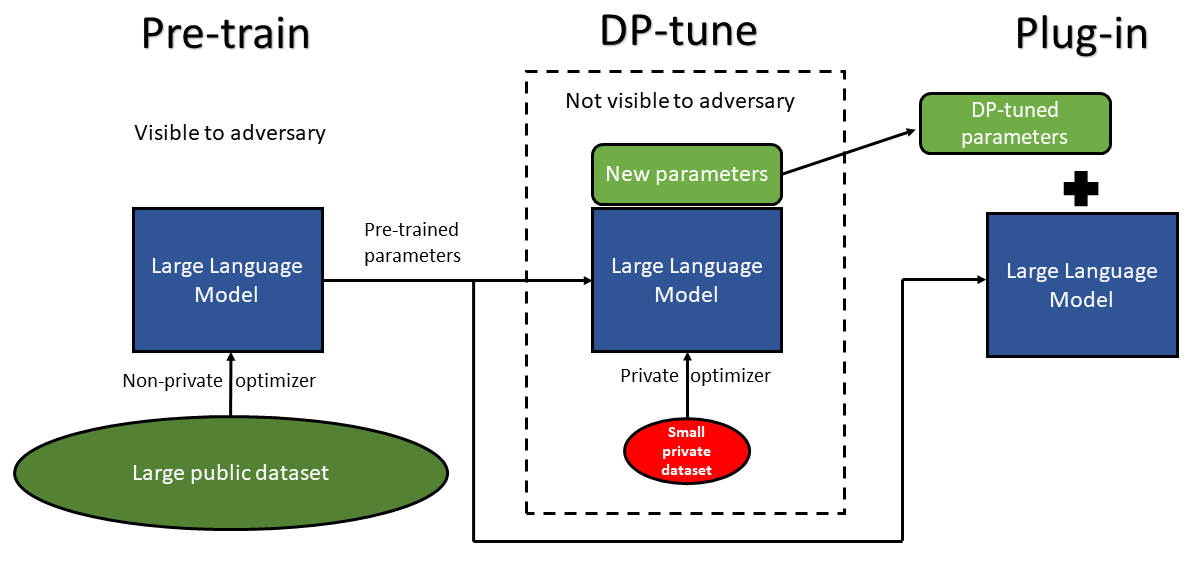}
  \caption{An illustration of our framework. First, the model is pre-trained on a large, public dataset. Next, new parameters are introduced and privately fine-tuned on a smaller, private, task-specific dataset. The original parameters are frozen during this process. Finally, the fine-tuned new parameters may be released publicly and plugged-in to the model for downstream tasks, while still preserving privacy of the private dataset.}
  \label{fig:framework}
\end{centering}
\end{figure}

\begin{table}
\centering
    \caption{Accuracy of fine-tuning for downstream tasks with RoBERTa-Large (in \%).
    Our results achieve accuracy comparable to full fine-tuning non-privately, while simultaneously guaranteeing differential privacy and modifying less than $1\%$ of the parameters. 
    We choose $\delta=$1e-5 for SST-2 and QNLI and $\delta=$1e-6 for MNLI and QQP due to their dataset sizes. Implementation details are in Section \ref{subsec:DP-RoBERTa}.
    }
\begin{tabular}{l|l|l|l|l|l|l}
        \hline
          Method & MNLI &  SST-2  &QQP &QNLI   &  Avg.  & Trained params\\
            \hline
              Non-private fine-tuning &     90.2 &  96.4  & 92.2  & 94.7 &  93.4 &  100\% \\\cline{1-7}
     Our results ($\epsilon = 6.7$) &       87.8      & 95.3   &  87.4    & 90.8  & 90.3  &   0.94\%   \\\hline 
        \end{tabular}
    \label{summary}
\end{table}

\begin{table}
\centering
    \caption{Fine-tuning GPT-2 models on the DART dataset. We observe that larger models have better utility, both in absolute numbers, and in terms of preserving non-private utility. DP parameters are ($\epsilon=6.8,\delta=$ 1e-5).} 
        \begin{tabular}{l|l|l|l}
        \hline
            Model & BLEU (DP) &  BLEU (non-private) & Drop due to privacy \\
            \hline
          GPT-2-Medium & 42.0 & 47.1 & 5.1 \\\cline{1-4}
          GPT-2-Large & 43.1 & 47.5 & 4.4 \\\cline{1-4}
          GPT-2-XL & 43.8 & 48.1  & 4.3  \\\cline{1-4}
            \hline 
        \end{tabular}
    \label{tbl:introtbl_dart}
\end{table}

Our primary contribution is to show that advanced parameter-efficient methods can lead to {\em simpler} and significantly improved algorithms for private fine-tuning.
Our framework is illustrated in Figure~\ref{fig:framework}. 
Our findings and contributions are summarized as follows:

\begin{itemize}
\item \textbf{State-of-the-art utility and privacy.}  Empirical evaluation of our algorithms reveals that they achieve state-of-the-art accuracy versus privacy tradeoffs, improving upon the previous best~\citep{yu2021large}. 
More importantly, for many fine-tuning tasks, the utility of models trained with DP approaches that of non-private models. 
For example, privately fine-tuning RoBERTa-Large on the MNLI data set~\citep{WilliamsNB18}, we achieve an accuracy of $87.8\%$ with a privacy budget of ($\epsilon = 6.7, \delta=$ 1e-6). 
Without privacy guarantees, RoBERTa-Large achieves an accuracy of $90.2\%$ (GPT-3 is known to achieve $91.7\%$ \citep{hu2021lora}); see Table \ref{summary} for a summary. 
We also explore private natural language generation tasks, fine-tuning GPT-2 models on the E2E dataset~\citep{NovikovaDR17}.
Again, the utility approaches non-private levels: we achieve a ROUGE-L score of 67.8 with GPT-2-Large and ($\epsilon = 6.0,\delta=$ 1e-5), compared to 72.0 without privacy.

\item \textbf{Larger models are better.}
Prior work has consistently shown that larger language models achieve better accuracy for downstream tasks.
Our results give evidence that this phenomenon extends to the private setting.
    For example, on the MNLI dataset, RoBERTa-Base achieves an accuracy of $83.5\%$ (versus $87.6\%$ non-privately, a drop of $4.1\%$) whereas RoBERTa-Large achieves an accuracy of $87.8\%$ (versus $90.2\%$ non-privately, a drop of $2.4\%$), both under a privacy budget of $(\epsilon = 6.7,\delta=$ 1e-6).
    Similarly, privately fine-tuning (using LoRA~\citep{hu2021lora}) on DART ($\epsilon = 6.8,\delta=$ 1e-5), GPT-2-Medium achieves a BLEU score of 42.0 (versus 47.1 non-privately, a drop of 5.1) while GPT-2-XL achieves a BLEU score of 43.8 (versus 48.1 non-privately, a drop of 4.3), see Table \ref{tbl:introtbl_dart}.
    Observe that utility improves with model size in two ways: both in terms of absolute numbers, as well as the drop incurred due to privacy.
    While the power of large models has been established in the non-private setting, we find this phenomenon quite surprising under DP.
  There is often a tension when choosing private model architectures: larger models may have higher capacity, but necessitate the introduction of more noise. 
    Consequently, smaller and simpler private models achieve the better accuracy in several settings~\citep{PapernotCSTE19, TramerB21}.
  In contrast, our experiments show that fine-tuning the biggest models achieves the best accuracy,\footnote{An alternative perspective is that what we currently think of as ``large'' language models are relatively small, and we are yet to reach the point where the benefits of model size on accuracy are outweighed by the drawbacks.} which we consider to be one of our main findings.

	\item \textbf{Simpler, sparser, and faster.} 
    Beyond accuracy concerns, DP requirements also lead to significant overheads in terms of computation and memory usage. 
    The large number of parameters contributes to the high cost of training LLMs, and things get worse under privacy, which has been documented to increase training time by up to two orders of magnitude~\citep{CarliniLEKS19,SubramaniVK21}.
    The parameter-efficient approaches we employ partially offset this issue: as we only update a small fraction of the total number of parameters, training becomes considerably more computationally and memory efficient.
    Furthermore, as in the non-private setting, this framework leads to a modular design, where a single large pre-trained model can be augmented with lightweight modifications for each individual downstream task. 

\end{itemize}

\medskip
To the best of our knowledge, we are the first to fine-tune GPT-2-XL using differential privacy, the largest model (with 1.5B parameters) trained thus far using DP. 
Given our state-of-the-art results for a variety of standard NLP tasks using advanced fine-tuning techniques, we believe that our paper will serve as a benchmark for further work in this direction. 
For example, the best average accuracy achieved by the prior work of \citet{yu2021large} on four standard NLP tasks in Table \ref{summary} is 
$83.9\%$ using $\epsilon = 8$ (and the same $\delta$ as in Table~\ref{summary}), whereas we can achieve an average accuracy of $90.3\%$ using $\epsilon = 6.7$ by a combination of better algorithms, larger models, and new privacy accounting techniques.

Finally, though recently considered elsewhere (see Section~\ref{sec:related}), we put further focus on the framing of public pre-training and private fine-tuning as an important conceptual direction in DP deep learning.

\section{Preliminaries and Prior Algorithm Baselines}
Recall the formal definition of differential privacy.
\begin{definition}[Differential Privacy (DP) \citep{DworkMNS06,DworkKMMN06}]
  A randomized algorithm $\mathcal{A}$ is  ($\epsilon$,$\delta$)-differentially private if for any two neighboring datasets $D$ and $D'$, which differ in exactly the data pertaining to a single user, and for all sets $\mathcal{S}$ of possible outputs: 
$
\textstyle{\Pr[\mathcal{A}(D) \in \mathcal{S}] \leq e^{\epsilon}\Pr[\mathcal{A}(D') \in \mathcal{S}] +\delta}
$.
\end{definition}
We review prior techniques for private fine-tuning.

\subsection{Full Fine-tuning via DPSGD} 
To train a machine learning model with privacy, the most popular algorithm is the celebrated DP stochastic gradient descent (DPSGD)~\citep{SongCS13, BassilyST14, AbadiCGMMTZ16}.
This optimization method serves as a drop-in replacement for SGD, augmenting it with the addition of per-example gradient clipping and Gaussian noise addition steps.
These two steps serve to limit and mask the contribution of a single example.
Two key points to note are that a) per-example gradient clipping incurs significant computational and memory overheads in most implementations, and b) noise introduced due to privacy grows as the square-root as the number of model parameters.
With this tool in place, the most basic fine-tuning strategy is to train all parameters using DPSGD.

\subsection{Reparametrized Gradient Perturbation}
To mitigate the limitations of DPSGD, a recent work of~\cite{yu2021large} introduced an elegant method called \emph{reparametrized gradient perturbation} (RGP). 
RGP exploits the implicit low-rank structure in the gradient updates of SGD to substantially improve upon DPSGD. 
Specifically, they reparametrize each layer's weight matrix $W$ into $LR + \tilde W$, where $L$ and $R$ are low-rank gradient-carrier matrices and $\tilde W$ is the residual weight.
The authors show that one can obtain a low-dimensional projection of $W$'s gradient by taking gradients only of the low-rank matrices $L$ and $R$ (and not the high-rank $\tilde W$).
Privacy is introduced by clipping and noising these low-dimensional gradients of $L$ and $R$.
While this low-dimensional projection loses some of the signal in $W$'s gradient, it turns out to contain enough to still achieve high accuracy.
At the same time, the low-dimensional gradients alleviate the aforementioned issues related to privatization, significantly reducing the memory consumption and noise introduced. Although RGP uses a low-rank update at each step, we empirically verify that its accumulated update is not of low stable rank and hence can not be compressed into small plug-in modules. Possible reasons include: 1) the low-rank subspaces of RGP are different  at different updates; 2) the accumulated update of RGP contains all the added noises, which are of high stable rank.


\newcommand{\WFT}{W_{\mathrm{FT}}}
\newcommand{\FT}{\mathrm{FT}}
\newcommand{\WPT}{W_{\mathrm{PT}}}
\newcommand{\PT}{\mathrm{PT}}

\section{Our Approach} \label{sec:newAlg}

\subsection{A Meta-framework}

We introduce our approach as a meta-framework for private deep learning, which abstracts the key principles of recent fine-tuning methods.

Suppose $f(\WPT;x)$ is a pre-trained model where $\WPT$ are the pre-trained weights and $x$ is any input. We create a new fine-tuning model
\begin{equation}
\label{eq:fine-tune}
 f_\FT(\WPT,\theta;x) 
\end{equation}
which incorporates additional trainable parameters $\theta$, where $\dim(\theta)\ll \dim(\WPT)$.
That is, the number of new parameters in $\theta$ is a small fraction of the original number of parameters in the pre-trained weights $\WPT$. 
Fine-tuning is done by running DPSGD on the additional parameters $\theta$, while freezing the weights of pre-trained model $\WPT$.
The new parameters are initialized to $\theta_0$ such that
\begin{equation}
\label{eq:fine-tune-initialization}
 f_\FT(\WPT,\theta_0;x) = f(\WPT;x).
\end{equation}
The initialization condition (\ref{eq:fine-tune-initialization}) is very important, as it ensures that fine-tuning starts at the pre-trained model and improves it by modifying the parameters $\theta$.
Most fine-tuning methods are additive and have the following special form:
\begin{equation}
\label{eq:fine-tune-additive}
 f_\FT(\WPT,\theta;x) = f(\WPT+\pi(\theta);x),
\end{equation}
i.e., they modify the pre-trained weights by adding a correction term $\pi(\theta)$ parametrized by $\theta$. 

Recent work in the non-private literature has described concrete instantiations of this framework~\citep{houlsby2019parameter, mahabadi2021compacter, hu2021lora}, which (crucially) are effective when $\dim(\theta)\ll \dim(\WPT)$.
In the non-private setting, such reparametrizations are useful for reducing the computation and memory required for fine-tuning, and enable lightweight and plug-in modifications to the base model for different downstream tasks.
At the same time, they maintain (or sometimes surpass) the accuracy achieved by full fine-tuning.


We give some intuition as to why parameter-efficient methods could be more effective for private fine-tuning especially when private datasets are small.
For simplicity, we assume that the fine-tuning method is additive as in (\ref{eq:fine-tune-additive}), such that the fine-tuned weights $\WFT=\WPT+\pi(\theta)$. 
We can imagine that $\WFT$ lies on a manifold passing through $\WPT$ of very small dimension (equal to the dimension of $\theta$) compared to the dimension of $\WPT$. 
Even if the parameters $\theta$ are very noisy due to the noise added during DPSGD, we will always stay in this manifold. 
In particular, we are not disturbing the pre-trained weights in most directions (those orthogonal to the manifold near $\WPT$). 
If we run DPSGD on all the weights instead, then we add noise in all directions, thus potentially unlearning the knowledge learned during pre-training, especially in low data regimes; see the discussion in Section \ref{sec:bad} for more on this.

Besides substantial gains in the accuracy, the above method of reparametrization has several other advantages:

\begin{itemize}
\item A single pre-trained model such as BERT or GPT is generally applied to hundreds of downstream tasks via fine-tuning.
Private fine-tuning using previous methods requires updating {\em all} parameters and storing a different copy of the fine-tuned model per task.
This creates substantial overheads for storing and deploying, and can be very expensive in practice. 
On the other hand, the reparametrization (\ref{eq:fine-tune}) means that we only need to store a single pre-trained model that can be shared across many downstream tasks. 
Each downstream task requires only a small number of new parameters that can be plugged in.

\item Differentially private training requires computing and storing per-example gradients, which increases the memory footprint. 
In our approach, however, learning is done in a much lower dimension, hence saving on the memory cost as compared to prior works.

\item Finally, we expect that (\ref{eq:fine-tune}) also gives a more communication-efficient method of fine-tuning in distributed settings such as federated learning, due to the significantly smaller number of parameters learned during fine-tuning. 

\end{itemize}

\subsection{Instantiating the Meta-framework}
In this section, we discuss a few ways to instantiate our meta-framework. 
This list is non-exhaustive, but covers the methods we employ in our experiments.

\subsubsection{Fine-tuning via Low-Rank Adaptation}

Low-Rank Adaptation (LoRA) \citep{hu2021lora} is an additive fine-tuning scheme as defined in (\ref{eq:fine-tune-additive}). 
For each dense weight matrix $W^i_{\PT}$ of size $a\times b$ in the pre-trained network, we add a low-rank correction term $L^iR^i$, i.e., 
\begin{flalign}
\label{eqn:LoRA}
W^i = W^i_\PT + L^iR^i,
\end{flalign} 
where $L^i\in \mathbb{R}^{a\times r}, R^i\in \mathbb{R}^{r\times b}$ are new trainable parameters. 
\citet{hu2021lora} apply this reparameterization only to the Transformer attention weights ($W_q, W_v$), and freeze all other weights (e.g., $W_k$ and $W_o$ and those in the feed-forward layers). 
The rank $r$ is typically chosen to be small, e.g., $r=4,16,64$. 
Since most parameters in Transformer architectures are dense weight matrices, choosing a small $r$ results in a nearly square-root reduction in the number of parameters. 

\subsubsection{Fine-tuning via Adapters}

\citet{houlsby2019parameter} propose adapter-based fine-tuning, in which we modify the architecture of the pre-trained model by adding new ``adapter'' layers after each attention and feed-forward layer.
Adapter layers are bottleneck layers with residual connections. 
Specifically, given an input $x$, an adapter layer $A$ performs
\begin{equation}
    \label{eqn:adapter}
A(x) = U(\mathrm{\tau}(D(x))) + x,
\end{equation}
where $U$ is an up-projection affine linear map, $D$ is a down-projection affine linear map, and $\tau$ is a non-linear activation function such as the Gaussian error Linear Unit (GeLU)~\citep{hendrycks2016gaussian}. 
If $x$ has dimension $d$, then $U\in \R^{d\times r},D\in \R^{r\times d}$ for some $r\ll d$. 
Thus, the number of introduced parameters is significantly less than the number of parameters in the pre-trained model.
When fine-tuning, the parameters of the original model are frozen, and only parameters of the adapter layers, as well as layer normalizations, are modified. 
Note that fine-tuning with adapters is not an additive fine-tuning framework as in (\ref{eq:fine-tune-additive}), but is captured by the broader framework in (\ref{eq:fine-tune}).

\subsubsection{Fine-tuning via Compacter}
The recent work of~\cite{mahabadi2021compacter} introduces Compacters (Compact adapters), a method which further improves the parameter efficiency of adapters. 
This is done by replacing the dense matrices in the up-projection $U$ and down-projection $D$ by tensor products of smaller matrices, thus reducing the number of trainable parameters.
Specifically, they replace the dense matrix $M_\ell$ in the adapter layer $\ell$ by a low-rank parameterized hypercomplex multiplication (LPHM) layer, i.e., each dense matrix $M_\ell \in \mathbb{R}^{a \times b}$ is expressed as 
\begin{equation}
    \label{eqn:compacter}
  M_\ell = \sum_{i=1}^n A_i \otimes \left(S_i^\ell  T_i^\ell\right)
\end{equation}
where $A_i\in \R^{n\times n}, S_i^\ell \in \R^{a/n \times k}, T_i^\ell \in \R^{k \times b/n}$ and $\otimes$ is the matrix Kronecker product. 
Note the matrices $A_i$ are not indexed by the layer $\ell$ because these matrices are shared among all the adapter layers. Since each adapter layers has two dense matrices (one for up-projection and one for down-projection), if there are $L$ adapter layers, this reduces the number of parameters from $L(2ab)$ to $L(2(a+b)k)+n^3$.
In practice, $a$ and $b$ are chosen to be either the model dimension $d$ or the intermediate representation dimension $r$ in the adapters, $n$ is typically chosen to be a small constant such as $n=2,4,8,12$ and $k$ is chosen to be $1$.

\subsubsection{Why Does Parameter-Efficient Tuning Work?}
Theoretical explanation of the success of parameter-efficient fine-tuning methods is an active area of research in deep learning.
Indeed, since trends have consistently shown that model accuracy increases with size, how can one achieve competitive accuracy while fine-tuning less than $1\%$ of the parameters?
One popular hypothesis is \emph{intrinsic dimensionality}~\citep{LiFLY18}, which posits that the minimum number of parameters needed to train a machine learning model may be much less than the total number of model parameters.
\citet{AghajanyanZG20} explore this hypothesis in the context of fine-tuning LLMs, showing that one can achieve most of their accuracy by training only a very small number of parameters (chosen via a random projection).
Perhaps surprisingly, they find that as {\em the model size increases, intrinsic dimension decreases}, in the limit exhibiting zero-shot learning.
While we did not explore this hypothesis in the context of DP due to computational restrictions, we believe it may be an interesting lens through which one can understand the effectiveness of private parameter-efficient fine-tuning.

\subsection{Comparision with Baseline Algorithms}
We highlight some key algorithmic differences between our proposed methods and the baselines of full fine-tuning and RGP.

\begin{itemize}
\item DPSGD and RGP both require updating all parameters of the pre-trained model, whereas our proposed methods update only a tiny fraction (between $0.05\%$ and $1\%$).
The rightmost columns of Tables~\ref{tbl:tbl_bert} and~\ref{tbl:tbl_Xbert} list the number of parameters trained by these algorithms. 

\begin{table}
\centering
    \caption{Memory and speed comparison for RoBERTa-Large. The rank is chosen as $r = 16$ for RGP and LoRA. 
    The speed is measured by the wall-clock time for training one epoch of the SST-2 dataset on a single Tesla V100 GPU with gradient accumulation for batch size 2000.}
\begin{tabular}{l|l|l}
        \hline
         Method          & Memory (GB) &  Speed (seconds per epoch)  \\
            \hline
              Full fine-tuning (DPSGD) & 27.9   & 715  \\\cline{1-3}
     RGP &  9.1     & 296 \\\cline{1-3}
DP LoRA   &   6.1    & 271 \\\hline 
        \end{tabular}
    \label{tbl:tbl_memory}
\end{table}
\item RGP performs a low-rank decomposition of weight matrices which is very similar to LoRA, though there are subtle differences.
    Recall that in RGP, at the beginning of each iteration $t$, the historical weight matrix $W_{t-1}$ is decomposed to find a low-rank product $LR$. 
The gradients computed on $L$ and $R$ are then projected back to the full parameter space to perform the descent step.
Hence, RGP does not keep the pre-trained weights frozen during the learning process.

LoRA can be viewed as a simplification of RGP.
    LoRA reparametrizes $W_{\text{FT}} := W_{\text{PT}} + LR$, where the pre-trained weight matrix $W_{\text{PT}}$ is frozen during training.
    Hence, compared to RGP, LoRA eliminates the decomposition and the projection to the full parameter space at each iteration, simplifying the implementation and reducing the running time and memory cost. 
This is summarized in Table \ref{tbl:tbl_memory}.
We observe that DP LoRA reduces the memory cost by about $33\%$ and the training speed by $8\%$.
As we will see, this simplification also results in improved utility.

\item Neither full fine-tuning nor RGP fall into our meta-framework described by (\ref{eq:fine-tune}).
Thus, if a pre-trained model is to be applied to several downstream tasks, one must store a separate set of weights for each task, incurring a significant memory cost and losing the plug-in functionality. 
In contrast, our methods are much more lightweight.
\end{itemize}

\section{Experiments}
\label{bertexps}
We experimentally evaluate our methods for DP fine-tuning to demonstrate their utility, privacy, and parameter-efficiency. 
We investigate both language understanding and text generation tasks, using RoBERTa and GPT-2 models, to establish that our techniques are applicable to a variety of tasks and model architectures.\footnote{Code for our experiments is available at \url{https://github.com/huseyinatahaninan/Differentially-Private-Fine-tuning-of-Language-Models}.}

\subsection{Fine-Tuning for Language Understanding Tasks}\label{subsec:DP-RoBERTa}

We first compare our methods with state-of-the-art fine-tuning algorithms using models from the BERT family, which was used in the prior work~\citep{yu2021large}. 
Specifically, we use RoBERTa models~\citep{LiuOGDJCLLZS19},\footnote{The model can be found at \url{https://github.com/pytorch/fairseq/tree/master/examples/roberta}.} which are pre-trained on public data collected from the web. 
RoBERTa-Base has 125M parameters and RoBERTa-Large has 355M parameters. 
We fine-tune the pre-trained models on four tasks: MNLI, QQP, QNLI and SST-2 from the GLUE benchmark~\citep{wang2018glue}, following~\citet{yu2021large}.

\textbf{Implementation Details:} 
For fine-tuning with adapters, we may choose the intermediate representation dimension $r$, shared across all adapter layers.
Similarly, for fine-tuning with Compacter, we can choose both the intermediate representation dimension $r$ and the Kronecker product kernel dimension $n$ in (\ref{eqn:compacter}). 
For LoRA fine-tuning, we add bottleneck branches for both the attention layers and the feedforward layers, which differs slightly from the addition of bottleneck branches for only the $W_q$ and $W_v$ matrices of the attention layers as done by~\cite{hu2021lora}. 
Given the same bottleneck representation dimension $r$ in (\ref{eqn:LoRA}), our new implementation uses twice as many trainable parameters as the original paper, and achieves some improvements for learning with DP.
We perform privacy accounting using the PRV Accountant from~\cite{gopi2021numerical}, which currently provides the tightest bounds.  

\begin{table}
\centering
    \caption{Accuracy for fine-tuning downstream tasks with RoBERTa-Base (in \%). The privacy parameters are $\epsilon=6.7$, and $\delta=$1e-5 for SST-2 and QNLI  and 1e-6 for MNLI and QQP. Bold indicates the best accuracy with DP. Numbers for non-private fine-tuning are from \citet{LiuOGDJCLLZS19}.} 

       \begin{tabular}{l|l|l|l|l|l|l|l}
        \hline
          \multicolumn{2}{c|}{Method}           & MNLI &  SST-2  &QQP &QNLI   &  Avg.  & Trained params\\
            \hline
             \multirow{2}{*}{  Full} & w/o DP   &     87.6       &   94.8    & 91.9  & 92.8 &  91.8 &  \multirow{2}{*}{100\%}\\\cline{2-7}
            &DP  &    53.1     &   82.6      & 74.4 & 63.9&  68.5  &  \\\cline{1-8}
            LoRA & w/o DP &    87.5         & 95.1   & 90.8  & 93.3  & 91.7   & 0.24\% \\\cline{1-8}  
            RGP\tablefootnote{We report RoBERTa-Base numbers from \url{https://github.com/dayu11/Differentially-Private-Deep-Learning}, by the authors of~\cite{yu2021large}.
            Though the paper itself only reports results on BERT-Base, we cite their paper to also reference the RoBERTa numbers.
            }  & DP &     80.1      &   91.6  & 85.5 & 87.2&86.1  & 100\% \\\cline{1-8}
           Adapter &DP &  83.4           &   \textbf{92.5} & 85.6 & \textbf{87.5}& \textbf{87.3} & 1.4\% ($r=48$)\\\cline{1-8}             
Compacter & DP&82.6 & 92.3 & 84.7 & 85.1 & 86.2 & 0.055\% ($r=96$, $n=8$) \\\cline{1-8}	LoRA  & DP&   \textbf{83.5}       & 92.2   & \textbf{85.7} & 87.3 &   87.2 & 0.94\% ($r=16$) 
            \\\hline 
    \end{tabular}
    \label{tbl:tbl_bert}
\end{table}

\begin{table}
\centering
    \caption{Accuracy for fine-tuning downstream tasks with RoBERTa-Large (in \%). The privacy parameters are $\epsilon=6.7$, and $\delta=$1e-5 for SST-2 and QNLI  and $\delta=$1e-6 for MNLI and QQP. Bold indicates the best accuracy with DP.
Numbers for non-private fine-tuning are from \citet{LiuOGDJCLLZS19}.
    } 
\begin{tabular}{l|l|l|l|l|l|l|l}
        \hline
          \multicolumn{2}{c|}{Method}           & MNLI &  SST-2  &QQP &QNLI   &  Avg.  & Trained params\\
            \hline
              Full &  w/o DP &     90.2 &  96.4  & 92.2  & 94.7 &  93.4 &  100\% \\\cline{1-8}
              LoRA&w/o DP &    90.6         &  96.2  & 91.6    & 94.9  &  93.3  & 0.23\% \\\cline{1-8}
     RGP&DP &   86.1         &   93.0 &  86.7 &  90.0 &  88.9  & 100\% \\\cline{1-8}
      Adapter  &DP &     87.7        & 93.9   &  86.3 & 90.7 &89.7  & 1.4\% ($r=48$) \\\cline{1-8}
Compacter &DP & 87.5 & 94.2 & 86.2 & 90.2 & 89.5  & 0.053\% ($r=96$, $n=8$) \\\cline{1-8}
LoRA & DP  &       \textbf{87.8}      & \textbf{95.3}   &  \textbf{87.4}    & \textbf{90.8}  & \textbf{90.3}  &   0.94\% ($r=16$)   \\\hline 
        \end{tabular}
    \label{tbl:tbl_Xbert}
\end{table}

\textbf{Hyperparameter choice:}  Given the large number of hyperparameter choices, e.g., the intermediate representation dimension, learning rate, weight decay, privacy delta, and model size, an exhaustive grid search over all hyperparameters is expensive, due to the model sizes.
Our hyperparameter choices are informed by prior work and are as follows.
For privacy parameters, we use $\delta=$ 1e-5 for SST-2 and QNLI  and $\delta=$ 1e-6 for QQP and MNLI due to their dataset sizes, and use noise multipliers $0.92, 0.83, 0.66$ and $0.65$ for SST-2, QNLI,  QQP and MNLI, respectively, which is the same as~\citet{yu2021large}\footnote{In Appendix~\ref{apdx:small_eps}, we evaluate the proposed framework  with various choices of privacy parameters.}.   The clipping threshold of per-example gradients is $10$ for all methods. 
For adapters and Compacter, we follow suggestions in the original papers and choose $r$ from a set $\{16, 48, 96\}$ and $n$ from a set $\{4, 8, 12\}$. 
For LoRA, we choose the best-performing rank $r$ from the set $\{4,16,48,64\}$.
The best performing hyperparameters are noted in Tables~\ref{tbl:tbl_bert} and~\ref{tbl:tbl_Xbert}.
We use batch size 2000 and train with half-precision  for 20 epochs. 
We use the optimizer AdamW~\citep{LoshchilovH19} with weight decay 1e-2 and search over four learning rates \{5e-4, 1e-3, 2e-3, 5e-3\}. In Appendix~\ref{apdx:heatmap}, we show the proposed algorithms perform well for a wide range of hyperparameters.

\textbf{Results:} We report the prediction accuracy on four tasks in Tables \ref{tbl:tbl_bert} and \ref{tbl:tbl_Xbert}. 
Our experiments using RoBERTa-Base serve as a direct comparison to the work of~\citet{yu2021large} who only trained the base model, whereas RoBERTa-Large experiments demonstrate the significance of using larger models.
We could not report the numbers for full fine-tuning using DPSGD on RoBERTa-Large due to running time and memory costs; see the discussion at the end of this section.
We summarize our key findings:
\begin{itemize}
\item On {\em all} datasets, our methods achieve the best accuracy while training a only tiny fraction of parameters; larger models give significant improvements.
\item Noticeable improvements in the privacy parameter $\epsilon$ versus~\cite{yu2021large} are primarily due to new privacy accountants based on Fourier-based numerical composition~\citep{koskela2020computing,koskela2021tight,gopi2021numerical}; we use the PRV Accountant from \citet{gopi2021numerical} since it is the most efficient.
\item Private adapters provide the best average performance for RoBERTa-Base, whereas LoRA outperforms all other methods for RoBERTa-Large. 
\end{itemize}

\begin{remark}
A concurrent work by Li {et al} \citep{LiTLH21} show that using a larger batch size and training with full-precision improves the performance of DPSGD, and obtains similar performance as our algorithm.
Thus, poor performance of DPSGD in our experiments is due to the suboptimal choice of hyperparameters and also due to precision issues, although we use same hyperparameters for all the algorithms.
We run new experiments with hyperparameters of  \citep{LiTLH21} in full precision mode, and get improvements around 1\% even for our algorithms. 
We report these findings in Appendix~\ref{apdx:large_bs}.
\end{remark}

\subsection{Fine-tuning for Natural Language Generation (NLG)}
Next, we study private fine-tuning for text generation problems using the GPT-2 series of models on the End-2-End (E2E) NLG challenge~\citep{NovikovaDR17} and DART~\citep{2021-dart},
two primary benchmarks used in recent works on non-private fine-tuning~\citep{hu2021lora, LiL21}.
We use GPT-2-Small (117M parameters), GPT-2-Medium (345M parameters), GPT-2-Large (774M parameters), and GPT-2-XL (1.5B parameters).\footnote{\url{https://huggingface.co/transformers/model_doc/gpt2.html}.}
To the best of our knowledge, we are the first to privately fine-tune for E2E-DART or fine-tune GPT-2-XL.
The purpose of this section is not to evaluate various fine-tuning algorithms, but to show that private fine-tuning is competitive with non-private fine-tuning for text generation problems.
Due to the high cost of training, we report experimental results only for fine-tuning (private and non-private) with LoRA. 
We think that all fine-tuning methods in this paper should achieve comparable accuracy.

\textbf{E2E NLG challenge:} The E2E dataset was introduced by \citet{NovikovaDR17}, and contains template-like information in the restaurant domain to be mapped to natural language with end-to-end training. 
The dataset consists of 42K training samples, 4.6K validation samples, and 4.6K test samples.

\textbf{DART:} DART was introduced as an open-domain data-to-text dataset by \citet{2021-dart}. 
The dataset consists of 62K training samples, 6.9K validation samples, and 12K test samples. 
In comparison to E2E, the dataset is larger and the task is more challenging. 

We use standard metrics such as BLEU, ROUGE-L, etc., used in \citep{hu2021lora} for measuring the quality of predictions.

\textbf{Hyperparameter choice:}
For LoRA, we choose the bottleneck rank $r = 4$ in (\ref{eqn:LoRA}) and fine-tune $W_q$ and $W_v$ matrices of the attention layers as in the original paper. 
We optimize using AdamW with learning rate 4e-4, weight decay 1e-2 and train our models for 20 epochs.
We use batch size 128 for the experiments on E2E and batch size 256 for the experiments on DART.
We take the gradient clipping parameter to be 1.0 and the noise multiplier to be 0.6 for the accountant in~\citet{gopi2021numerical}, achieving $\epsilon = 6.0,\delta=$1e-5 on E2E and $\epsilon = 6.8,\delta=$1e-5 on DART.

\textbf{Results:} The results of our experiments are summarized in the Table \ref{tbl:tbl_e2e} and \ref{tbl:tbl_dart}, which reiterate the main themes of our work:
private fine-tuning with parameter-efficient approaches perform close to their non-private counterparts and show consistent improvement in the utility as model size increases.
Note that on E2E dataset, although the validation perplexity improves as the model becomes larger, the metrics seem to saturate going from large to XL for both private and non-private cases. 
On the other hand, for DART dataset both validation perplexity and the metric improve as the model size increases.

\begin{table}
\centering
    \caption{Metrics on the E2E NLG task. Non-DP results from~\citet{hu2021lora}, except for GPT-2-XL, which was not reported in the paper. We ran GPT-2-XL with hyperparameters presented in~\citet{hu2021lora}. Bold indicates the best accuracy with DP. DP parameters are ($\epsilon=6.0,\delta=$ 1e-5). Val perp stands for validation perplexity.} 
        \begin{tabular}{l|l|l|l|l|l|l}
        \hline
            Method & Val perp & BLEU &  NIST & MET & ROUGE-L   &  CIDEr  \\
            \hline
		GPT-2-Small + DP  & 4.51 &  63.8 & 7.19 & 39.5 & 67.5 & 1.87 \\\cline{1-7}
		GPT-2-Medium + DP  & 4.02 & 65.5 & 8.45 & 42.7 & 67.9 & 2.23  \\\cline{1-7}
          GPT-2-Large + DP  & 3.87 & \textbf{66.7} & \textbf{8.63} & \textbf{44.0} & 67.8 & \textbf{2.33} \\\cline{1-7}
          GPT-2-XL + DP & \textbf{3.79}  & 66.1 & 8.53 & 43.0 & \textbf{68.1} & 2.28 \\\cline{1-7}
    GPT-2-Medium & 3.19 & 70.4  & 8.85  &  46.8 & 71.8 &  2.53  \\\cline{1-7}
		GPT-2-Large  & 3.06 & 70.4  & 8.89  &  46.8 & 72.0 &  2.47  \\\cline{1-7}
		GPT-2-XL  & 3.01 & 69.4 & 8.78 & 46.2 & 71.5 & 2.49 \\\cline{1-7}
            \hline 
        \end{tabular}
    \label{tbl:tbl_e2e}
\end{table}

\begin{table}
\centering
    \caption{Metrics on the DART dataset. Non-DP results from~\citet{hu2021lora}, except for GPT-2-XL, which was not reported in the paper. We ran GPT-2-XL with hyperparameters presented in~\citet{hu2021lora}. Bold indicates the best accuracy with DP. DP parameters are ($\epsilon=6.8,\delta=$ 1e-5). Val perp stands for validation perplexity. Unlike all other metrics, the lower the TER metric is the better for the performance of the model.} 
        \begin{tabular}{l|l|l|l|l}
        \hline
            Method     & Val perp  & BLEU &  MET & TER  \\
            \hline
		GPT-2-Small + DP  & 3.82 & 38.5 & 0.34 & 0.53 \\\cline{1-5}
		GPT-2-Medium + DP & 3.30  & 42.0 & 0.36 & 0.51 \\\cline{1-5}
          GPT-2-Large + DP & 3.10 & 43.1 & 0.36 & \textbf{0.5} \\\cline{1-5}
          GPT-2-XL + DP & \textbf{3.00} & \textbf{43.8} & \textbf{0.37} & \textbf{0.5} \\\cline{1-5}
        		GPT-2-Medium & 2.67 & 47.1 & 0.39 & 0.46 \\\cline{1-5}
		GPT-2-Large & 2.89 & 47.5 & 0.39 & 0.45 \\\cline{1-5}
		GPT-2-XL & 2.83 & 48.1 & 0.39 & 0.46 \\\cline{1-5}
            \hline 
        \end{tabular}
    \label{tbl:tbl_dart}
\end{table}

\subsection{How Bad is DPSGD?} 
\label{sec:bad}
While our experiments indicate that full fine-tuning  does not achieve competitive performance, there could be a choice of hyperparameters that improves upon the reported numbers, e.g., ``mega'' batch sizes (in the millions) in \citet{AnilGGKM21,LiTLH21}. We note that our main message is that one does not need to fine-tune all parameters to achieve the best accuracy. Nevertheless, it is interesting to wonder if full fine-tuning with DPSGD can match the accuracy of parameter-efficient methods. A positive answer would imply that private and non-private fine-tuning conceptually mirror each other. 

\textbf{Update:} A concurrent work by Li et al  \citep{LiTLH21} answer our question and show that full fine-tuning via DPSGD also achieves same performance as the parameter efficient methods used in this paper. 
These two works raise an intriguing theoretical question: Why does magnitude of the noise added by DPSGD has no impact on the accuracy vs privacy tradeoffs when fine-tuning? Explaining this  phenomenon would deepen our understanding of private learning.
\section{Related Work}
\label{sec:related}
\paragraph{More on DP learning:} Some work studies private language models on more traditional architectures such as LSTMs~\citep{HochreiterS97}, either training with DPSGD~\citep{McMahanRTZ18,CarliniLEKS19} or related heuristics~\citep{RamaswamyTMAMB20}.
Though pre-training on public data is suggested~\citep{McMahanRTZ18}, public data appears to only be used in one of these works for honest hyperparameter selection~\citep{RamaswamyTMAMB20}.
A few more recent works consider training LLMs with DP. 
\citet{AnilGGKM21} privately train BERT-Large from scratch, compared to our work which focuses on private fine-tuning.
~\citep{HooryFTCELNSBHM21, BasuRNMSM21} perform private full fine-tuning of BERT models.
\citet{HooryFTCELNSBHM21} achieve accuracy which is comparable to the non-private model, but additionally supplement the public pre-training data with additional domain-relevant material, while we use off-the-shelf pre-trained models.
\cite{BasuRNMSM21} observe significant drops in utility, compared to our parameter-efficient methods which do not.
While~\citet{KerriganST20} consider public pre-training and private fine-tuning, their experiments are on much smaller architectures (i.e., feedforward networks with three hidden layers).
A simultaneous work of~\cite{GinartVZG22} investigates private \emph{prediction} (rather than learning) for next-token prediction. 
A subsequent work by~\cite{SengeIH21} also investigates the effect of private fine-tuning on various NLP tasks. 

Our investigation fits more broadly into a line of work employing public data for private data analysis.
Some works on image classification consider pre-training on a large public dataset and fine-tuning on a smaller private dataset~\citep{AbadiCGMMTZ16, PapernotCSTE19, TramerB21, LuoWAF21}.
In particular, \citet{LuoWAF21} investigate the role of parameter efficiency in private fine-tuning ResNet models, and propose strategies to choose which parameters to fine-tune.
One line of work uses unlabeled public data to train a student model~\citep{PapernotAEGT17, PapernotSMRTE18, BassilyTT18}, including one work simultaneous to our own for natural language generation~\citet{tian2022seqpate}.
Another recent idea uses a small amount of public data to identify a lower-dimensional subspace of the gradients in which to perform private descent~\citep{ZhouWB21, YuZCL21, KairouzRRT21}.
A simultaneous work of~\citet{AmidGMRSSSTT21} uses public data in the mirror map for a private mirror descent algorithm.
Finally, other works (both theoretical and experimental) investigate the role of public data in private query release, synthetic data generation, and prediction~\citep{JiE13, BeimelNS16, AlonBM19, NandiB20, BassilyCMNUW20, BassilyMN20, LiuVSUW21}.

In a concurrent work, \citet{LiTLH21} also investigate DP fine-tuning of LLMs. In several cases, their results demonstrate qualitatively similar findings as ours.
While our experiments focus primarily on parameter-efficient fine-tuning methods, interestingly, they show that private full fine-tuning can also achieve comparable utility if the experimental setup is configured properly, e.g., using suitable hyperparameters. In Appendix~\ref{apdx:large_bs}, we run experiments under the setup in \citet{LiTLH21}. We show their setup can also improve the performance of our methods.

\paragraph{More on Fine-tuning:} There exist other parameter-efficient tuning methods which we did not evaluate in our work. 
Some of these include random subspace projection (exploiting intrinsic dimensionality~\citep{LiFLY18, AghajanyanZG20}), prefix and prompt tuning~\citep{LiL21, LesterAC21}, tuning only biases~\citep{CaiGZH20, BenZakenRG21}, and other architecture variants including Adapters~\citep{PfeifferKRCG21, RuckleGGBPRG20}.
Other works investigate lightweight methods for adapting language models to different tasks (e.g.,~\citet{DathathriMLHFMYL20})
An interesting direction for future work is to see whether parameter-efficient tuning approaches specifically designed for the private setting can achieve higher utility.
We also mention zero-shot learning, in which no task-specific dataset is required and thus perfect privacy is achieved.
Currently, zero-shot approaches achieve low utility compared to fine-tuning, though it is possible that future models may narrow this gap.

\section{Conclusion}
So far, DP deep learning has focused on training models from scratch.
The spectacular success of transfer learning in real-world applications, however, shows that private fine-tuning is an equally pertinent problem to study and deserves more attention.
We show that by combining recent advances in NLP, parameter-efficiency, privacy accounting, and using larger models, one can privately fine-tune models whose utility approaches that of non-private models. 
We hope our work inspires more study on the core problem of private fine-tuning, which we believe to be a central direction for research in private machine learning, leading to more interaction between the LLM and DP communities.

\section*{Acknowledgments}
The authors would like to thank Rabeeh Karimi Mahabadi for sharing hyperparameters based on experiments in~\cite{mahabadi2021compacter} and 
Xuechen Li for sharing the experimental setup and many suggestions about mixed-precision training.
Janardhan Kulkarni would like to thank Edward Hu for sharing ideas on fine-tuning.

\bibliography{biblio}
\bibliographystyle{plainnat}

\appendix

\newpage
\section{Experiments with Different Privacy Parameters}
\label{apdx:small_eps}

Now we test our framework under different privacy constraints. Specifically, we run LoRA on the language understanding tasks with various choices of  privacy parameters $\epsilon$ and $\delta$. We consider both RoBERTa-Base and RoBERTa-Large. 

For the RoBERTa-Large model, we set $\epsilon=1$ and $3$ with $\delta$ being the same as those in Section~\ref{bertexps}. We use the PRV accountant \citep{gopi2021numerical}. After getting the noise multipliers, we also reduce the value of $\delta$ and report the corresponding value of $\epsilon$.  The hyperparameters are the same as those in Section~\ref{bertexps}.  We run experiments on all four tasks, i.e.,  MNLI ($n\sim$ 392k), QQP ($n\sim$ 364k), QNLI ($n\sim$ 104k), and SST-2 ($n\sim$ 67k).  We report the results in Table~\ref{tbl:tbl_Xbert_smalleps}. The performance of our framework is decent even with very tight privacy budgets. For instance, with $\epsilon<2$ and $\delta=1/1000n$, the accuracy gap between the non-private baseline is only 3.8 for MNLI and  2.1 for SST-2. 



\begin{table}
\renewcommand{\arraystretch}{1.25}
\centering
    \caption{Test accuracy for fine-tuning RoBERTa-Large with different privacy parameters. The number of training samples is denoted by $n$. The values of $\sigma$ are noise multipliers. Numbers in the brackets are the changes compared to the results in Table~\ref{tbl:tbl_Xbert} ($\epsilon=6.7$, $\delta=\Theta(1/n)$).  } 

       \begin{tabular}{l|l|l|l|l|l||l}
        \hline
        \hline
          Taks  & $\sigma$     &  $\delta=1/n$ & $\delta=1/10n$ &$\delta=1/100n$ & $\delta=1/1000n$   & Accuracy  (in \%) \\
            \hline
            MNLI & 1.88  &  $\epsilon=1$  &  $\epsilon=1.35$   &  $\epsilon=1.49$    & $\epsilon=1.61$ &  86.8 (-1.0\%)  \\\hline
            
            QQP  & 1.88 &  $\epsilon=1$  &  $\epsilon=1.40$   &  $\epsilon=1.54$    & $\epsilon=1.67$ &  85.2 (-2.2\%)  \\\hline            
            
            QNLI & 3.01 &  $\epsilon=1$  &  $\epsilon=1.48$   &  $\epsilon=1.64$    & $\epsilon=1.79$ & 88.0 (-2.8\%)   \\\hline 

             SST-2 & 3.63 &  $\epsilon=1$  &  $\epsilon=1.47$   &  $\epsilon=1.64$    & $\epsilon=1.80$ & 93.1  (-2.2\%) \\\hline \hline
 
            MNLI & 0.91 &  $\epsilon=3$  &  $\epsilon=4.12$   &  $\epsilon=4.51$    & $\epsilon=4.89$ &   87.4 (-0.4\%)  \\\hline

            QQP  &0.93 &  $\epsilon=3$  &  $\epsilon=4.10$   &  $\epsilon=4.49$    & $\epsilon=4.86$ &  86.8  (-0.6\%)  \\\hline

            QNLI  &1.29 &  $\epsilon=3$  &  $\epsilon=4.45$   &  $\epsilon=4.90$    & $\epsilon=5.33$ &  89.9 (-0.9\%)  \\\hline

              SST-2 &  1.52 &  $\epsilon=3$  &  $\epsilon=4.37$   &  $\epsilon=4.83$    & $\epsilon=5.25$ & 94.1  (-1.2\%) \\\hline
    \end{tabular}
    \label{tbl:tbl_Xbert_smalleps}
\end{table}

\begin{figure} [h]
\centering
\begin{subfigure}{.5\textwidth}
  \centering
  \includegraphics[width=.9\linewidth]{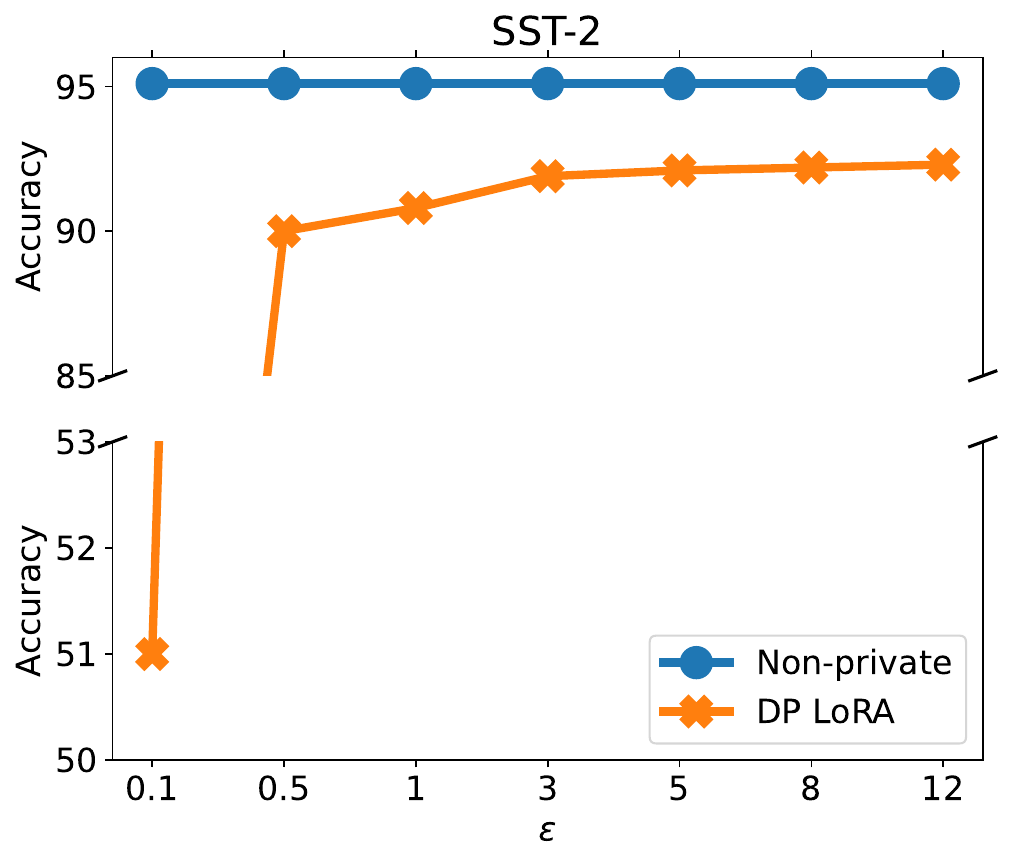}
  \label{fig:sub1}
\end{subfigure}%
\begin{subfigure}{.5\textwidth}
  \centering
  \includegraphics[width=.9\linewidth]{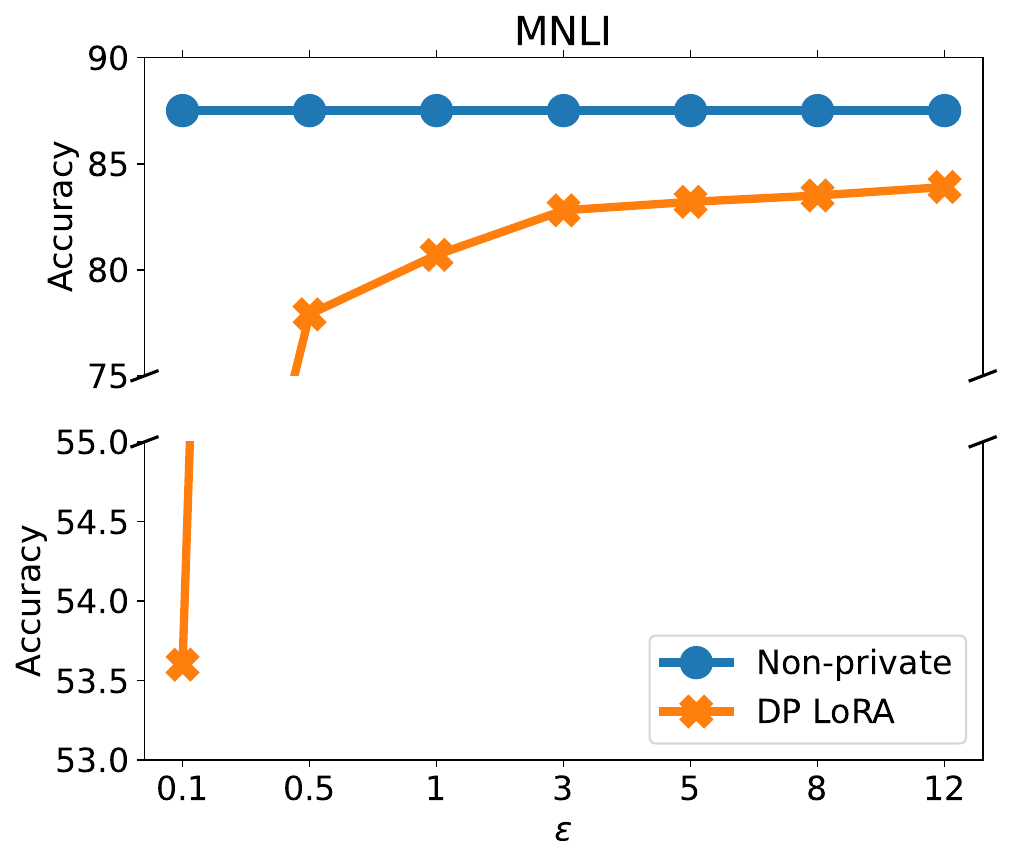}
  \label{fig:sub2}
\end{subfigure}
\caption{Test accuracy (in \%) of fine-tuning the RoBERTa-Base model on MNLI and SST-2 with various choices of $\epsilon$.}
\label{fig:fig_bert_smalleps}
\end{figure}

For the RoBERTa-Base model, we try various choices of $\epsilon$. The values of $\epsilon$ are chosen from $[0.1, 0.5, 1, 3, 5, 8, 12]$. All other settings are the same as those in Section~\ref{bertexps}. We run experiments on the MNLI and SST-2 datasets. The results are presented in Figure~\ref{fig:fig_bert_smalleps}. Our framework performs well for a wide range of  $\epsilon$. We note that our algorithm achieves meaningful accuracy even for very tight privacy parameters $\epsilon=0.5$ and $1$. Such values of $\epsilon$ are rarely explored when training deep models with differential privacy.

\section{On the Influence of Hyperparameters}
\label{apdx:heatmap}

\begin{figure}
\begin{centering}
\includegraphics[width=0.95\textwidth]{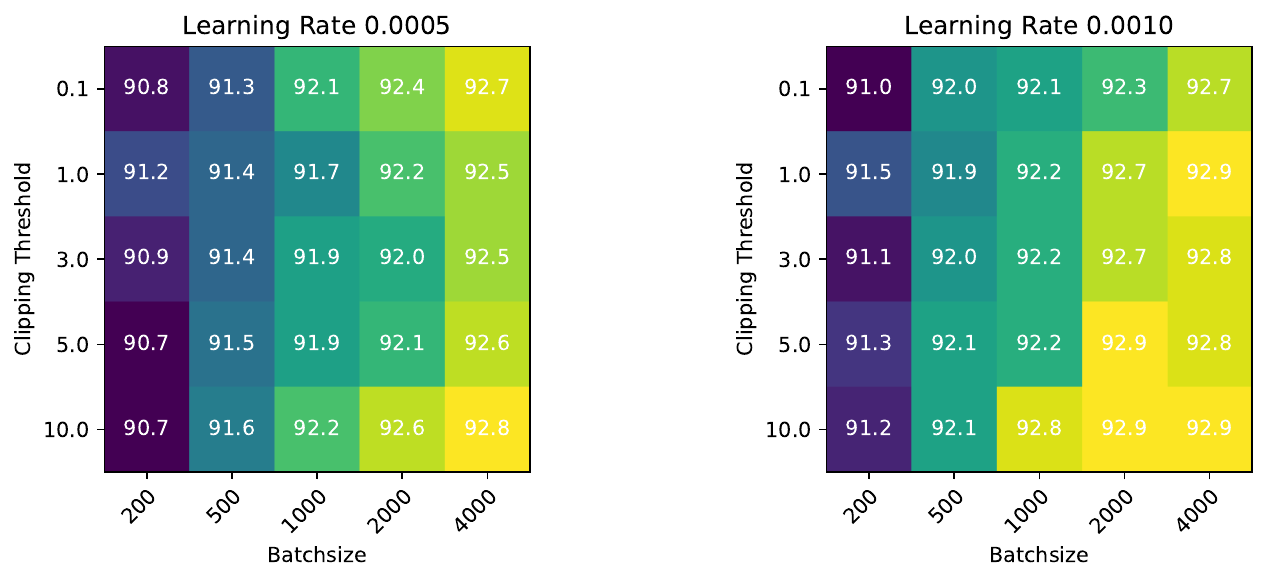}
  \caption{Test accuracy (in \%) of fine-tuning RoBERTa-Base with differentially private LoRA on the SST-2 dataset. Our algorithm performs well on a wide range of hyperparameters.}
  \label{fig:fig_bert_heatmap}
\end{centering}
\end{figure}

Here we demonstrate that our algorithms perform well for a wide range of hyperparameters. We study two hyperparameters that are directly related to the variance of noise: clipping threshold and batchsize. The clipping threshold is chosen from $[0.1, 1.0, 3.0, 5.0, 10.0]$ and the batchsize is chosen from $[200, 500, 1000, 2000, 4000]$. We note that we keep the number of updates the same as that in Section~\ref{bertexps} when the batchsize is changed. We fine-tune the RoBERTa-Base model with differentially private LoRA ($r=16$)  on the SST-2 dataset. The results are presented in Figure~\ref{fig:fig_bert_heatmap}. DP LoRA performs well for all the hyperparameters considered. The gap between the best accuracy and the worst accuracy is only 2\%.

\section{Fine-Tuning for Language Understanding Tasks with Large Batch Size and Full-Precision}
\label{apdx:large_bs}

\citet{LiTLH21} show the performance of fine-tuning the full model is sensitive to the choice of hyperparameters. They give a configuration which can significantly improve the performance of full fine-tuning.  In this section, we re-evaluate the tasks in Table~\ref{tbl:tbl_bert} and~\ref{tbl:tbl_Xbert} under the configuration in \citet{LiTLH21}. 

\begin{table}
\centering
    \caption{Accuracy for fine-tuning downstream tasks with RoBERTa-Base (in \%). Experiments are run with full-precision. We also scale up the batch size according to the dataset size compared to SST-2. The privacy parameters are $\epsilon=6.7$, and $\delta=$1e-5 for SST-2 and QNLI  and 1e-6 for MNLI and QQP.} 

     \begin{tabular}{l|l|l|l|l|l|l}
        \hline
          \multicolumn{2}{c|}{Method}           & MNLI &  SST-2  &QQP &QNLI   & Average Accuracy  \\
            \hline
             \multirow{2}{*}{  Full} & w/o DP   &     87.6       &   94.8    & 91.9  & 92.8 &  91.8 \\\cline{2-7}
            &DP  &  83.2       &    85.9    & 86.2  & 84.8 &    85.0 \\\cline{1-7}
           Adapter &DP &   \textbf{84.6}        & \textbf{92.9}   & \textbf{87.4}  & \textbf{89.2} & \textbf{88.5}  \\\cline{1-7}             
        	LoRA  & DP&    84.5     & 92.7   & 87.1  & 88.3  & 88.2   
            \\\hline 
        \end{tabular}
    \label{tbl:bert_largebs}
\end{table}

\begin{table}
\centering
    \caption{Accuracy for fine-tuning downstream tasks with RoBERTa-Large (in \%). Experiments are run with full-precision. We also scale up the batch size according to the dataset size compared to SST-2. The privacy parameters are $\epsilon=6.7$, and $\delta=$1e-5 for SST-2 and QNLI  and $\delta=$1e-6 for MNLI and QQP.
    } 
     \begin{tabular}{l|l|l|l|l|l|l}
        \hline
          \multicolumn{2}{c|}{Method}           & MNLI &  SST-2  &QQP &QNLI   &  Average Accuracy  \\
            \hline
             \multirow{2}{*}{  Full} & w/o DP   &     90.2       &   96.4    & 92.2  & 94.7 &  93.4 \\\cline{2-7}
            &DP  &     86.4    &    90.9    & 87.5  & 89.4 & 88.6     \\\cline{1-7}
           Adapter &DP &     88.6       &  94.5   & 87.8  & 91.6 & 90.6  \\\cline{1-7}             
        	LoRA  & DP&    \textbf{89.0}     & \textbf{95.3}    & \textbf{88.4}  & \textbf{92.4}  & \textbf{91.3} 
            \\\hline 
        \end{tabular}
    \label{tbl:Xbert_largebs}
\end{table}

The configuration in \citet{LiTLH21} has two differences compared to that in Section~\ref{bertexps}. The first difference is \citet{LiTLH21} run experiments with full-precision while the experiments in Section~\ref{bertexps} use half-precision. Using half-precision is a common approach to speed up NLP experiments \citep{ott2018scaling}. However, half-precision may incur underflow issue which impacts the model performance \citep{micikevicius2017mixed}. The second difference is they use larger batch size for larger datasets. For example, the batch size for MNLI is roughly six times larger than the batch size for SST-2 in \citet{LiTLH21}. In Section~\ref{bertexps}, we use the same batch size for all datasets.

We follow the above setup and re-evaluate DP-LoRA and DP-Adapter. The results are in Table~\ref{tbl:bert_largebs} and~\ref{tbl:Xbert_largebs}. The results of full fine-tuning with differential privacy are directly adopted from \citet{LiTLH21}. The configuration in \citet{LiTLH21} further improves the strong results in Table~\ref{tbl:tbl_bert} and~\ref{tbl:tbl_Xbert}. For example, we achieve 89.0\% accuracy on the MNLI dataset, which is only 1.2\% lower than the accuracy without DP constraint. Moreover, the benefit of the proposed framework over full fine-tuning is still clear. The average accuracy of the proposed algorithms is $\sim$3\% higher than that of full fine-tuning.

\end{document}